\pdfoutput=1
\documentclass{article}
\PassOptionsToPackage{table}{xcolor}
\usepackage{iclr2027_conference,times}
\usepackage[T1]{fontenc}
\usepackage[utf8]{inputenc}
\usepackage{amsmath,amssymb,graphicx,booktabs,array}
\usepackage{algorithm,algpseudocode}
\usepackage{xcolor}
\usepackage{hyperref}
\usepackage{xurl}
\hypersetup{colorlinks=true,linkcolor=blue,citecolor=blue,urlcolor=blue}
\title{EvoWeave: Composing Reusable Agent\\Experience through Operation Level Graphs}
\author{Guanhua Zhu$^{1}$, Bo Shen$^{2}$, Lei Zhang$^{1}$, Fadi Deek$^{1}$, \\
{\bf Zhi Wei$^{3}$, Mengnan Du$^{4,*}$, Chenxi Yuan$^{1,*}$} \\
{\normalfont $^{1}$Department of Informatics, New Jersey Institute of Technology} \\
{\normalfont $^{2}$Department of Mechanical and Industrial Engineering, New Jersey Institute of Technology} \\
{\normalfont $^{3}$Department of Computer Science, New Jersey Institute of Technology} \\
{\normalfont $^{4}$School of Artificial Intelligence, The Chinese University of Hong Kong, Shenzhen} \\
{\normalfont\texttt{\{gz72,bo.shen,lei.zhang,fadi.deek,zhi.wei,chenxi.yuan\}@njit.edu}} \\
{\normalfont\texttt{mengnandu@cuhk.edu.cn}} \\
{\normalfont $^{*}$Corresponding authors}}
\iclrfinalcopy
\newcommand{\method}{EvoWeave}
\newcommand{\MainNormal}{51.91}
\newcommand{\NoNormal}{44.92}
\newcommand{\MainChallenge}{27.99}
\newcommand{\NoChallenge}{23.01}
\newcommand{\MainWeb}{44.90}
\newcommand{\NoWeb}{38.89}
\newcommand{\MainAlf}{70.96}
\newcommand{\NoAlf}{62.44}
\newcommand{\MainScience}{66.97}
\newcommand{\NoScience}{56.93}
\newcommand{\GainNormal}{6.99}
\newcommand{\BestNormalGain}{2.01}
\newcommand{\GainChallenge}{4.99}
\newcommand{\BestChallengeGain}{1.66}
\newcommand{\MainWins}{16}

\newcommand{\AblB}{29.79}
\newcommand{\AblC}{31.39}
\newcommand{\AblD}{33.47}
\newcommand{\AblE}{32.82}
\newcommand{\AblF}{34.86}
\newcommand{\AblCDGain}{2.08}
\newcommand{\AblDFGain}{1.39}
\newcommand{\AblCEGain}{1.44}
\newcommand{\AblFEGap}{2.04}
\newcommand{\GraphStates}{27.0}
\newcommand{\GraphOperations}{157.6}
\newcommand{\GraphCodeNodes}{185.2}
\newcommand{\GraphCodeLinks}{157.6}
\newcommand{\GraphSkillNodes}{82.2}
\newcommand{\GraphSkillLinks}{362.0}
\newcommand{\GraphRoleNodes}{6.8}
\newcommand{\GraphRoleLinks}{99.4}
\newcommand{\InjectOurs}{6,155.46}
\newcommand{\InjectFlat}{26,300.80}
\newcommand{\InjectAWM}{19,890.00}
\newcommand{\InjectRB}{11,872.00}
\newcommand{\SaveFlat}{76.60}
\newcommand{\SaveAWM}{69.05}
\newcommand{\SaveRB}{48.15}
\newcommand{\EnterSaving}{0.99}
\newcommand{\RoundsOurs}{15.94}
\newcommand{\SecondsOurs}{10.44}
\newcommand{\KVOurs}{45}
\newcommand{\CatChallengeMin}{17.01}
\newcommand{\CatChallengeMax}{36.83}
\newcommand{\CatAlfMin}{61.57}
\newcommand{\CatAlfMax}{79.26}
\newcommand{\CatScienceMin}{45.63}
\newcommand{\CatScienceMax}{88.41}
\newcommand{\OverlapNormal}{48.5}
\newcommand{\ReferenceNormal}{49.2}
\newcommand{\OverlapChallenge}{57.3}
\newcommand{\ReferenceChallenge}{58.8}
\newcommand{\OverlapWeb}{35.3}
\newcommand{\ReferenceWeb}{36.7}
\newcommand{\OverlapAlf}{96.5}
\newcommand{\ReferenceAlf}{96.9}
\newcommand{\OverlapScience}{90.2}
\newcommand{\ReferenceScience}{92.1}
\newcommand{\WebDiagOursMax}{1.26}
\newcommand{\WebDiagBaseMin}{1.41}
\newcommand{\WebDiagMin}{1.23}
\newcommand{\WebDiagMax}{1.69}
\newcommand{\AlfDiagMin}{-0.060}
\newcommand{\AlfDiagMax}{0.074}

\algrenewcommand\algorithmicrequire{Input:}
\algrenewcommand\algorithmicfor{for}
\algrenewcommand\algorithmicdo{do}
\algrenewcommand\algorithmicif{if}
\algrenewcommand\algorithmicthen{then}
\algrenewcommand\algorithmicelse{else}
\algrenewcommand\algorithmicend{end}
\algrenewcommand\algorithmicreturn{return}
\makeatletter
\renewcommand{\floatc@ruled}[2]{{\normalfont #1 #2}\par}
\makeatother
\begin{document}
\maketitle
\lhead{Preprint.}
\begin{abstract}
Different tasks often require interactive agents to perform similar operations at intermediate steps. Retrieving an entire past workflow can introduce irrelevant steps, whereas retrieving isolated operations can discard the dependencies needed to reuse them. We introduce EvoWeave, a framework for experience memory that represents abstract states as nodes and reusable operations as directed edges with input and output contracts, provenance, and capability references. A planner decomposes the current task into a dependency graph; operation retrieval and bounded directed search then assemble relevant experience fragments. An execution agent can inspect these fragments and their referenced capabilities, while reflection separates concrete execution evidence from reusable abstractions. We study the framework across four benchmarks, five evaluation settings, and three language model backbones. Across backbones, AppWorld Normal and Challenge task goal completion average \MainNormal{} and \MainChallenge{}, respectively, exceeding the baseline without memory by \GainNormal{} and \GainChallenge{} percentage points. An AppWorld ablation associates directed connections with an additional improvement of \AblCDGain{} points over independent operation cards. The results motivate composition of reusable operations while exposing tradeoffs in retrieval overhead and conditions for valid reuse.
\end{abstract}

\section{Introduction}
Language model agents solve interactive tasks by interleaving reasoning with actions and environmental feedback \citep{yao2022react}. Although task goals vary, many constituent operations may recur across tasks: locating a contact, filtering records, reading an attribute, modifying an object, or submitting a result. An agent that has already performed these operations should be able to reuse that experience when pursuing a different goal. Approaches based on experience therefore reuse insights or workflows that an agent has learned from prior interactions \citep{zhao2024expel,wang2024agent,ouyang2026reasoningbank}, making the organization of experience a central design choice.

\begin{figure}[t]
\centering
\includegraphics[width=\linewidth]{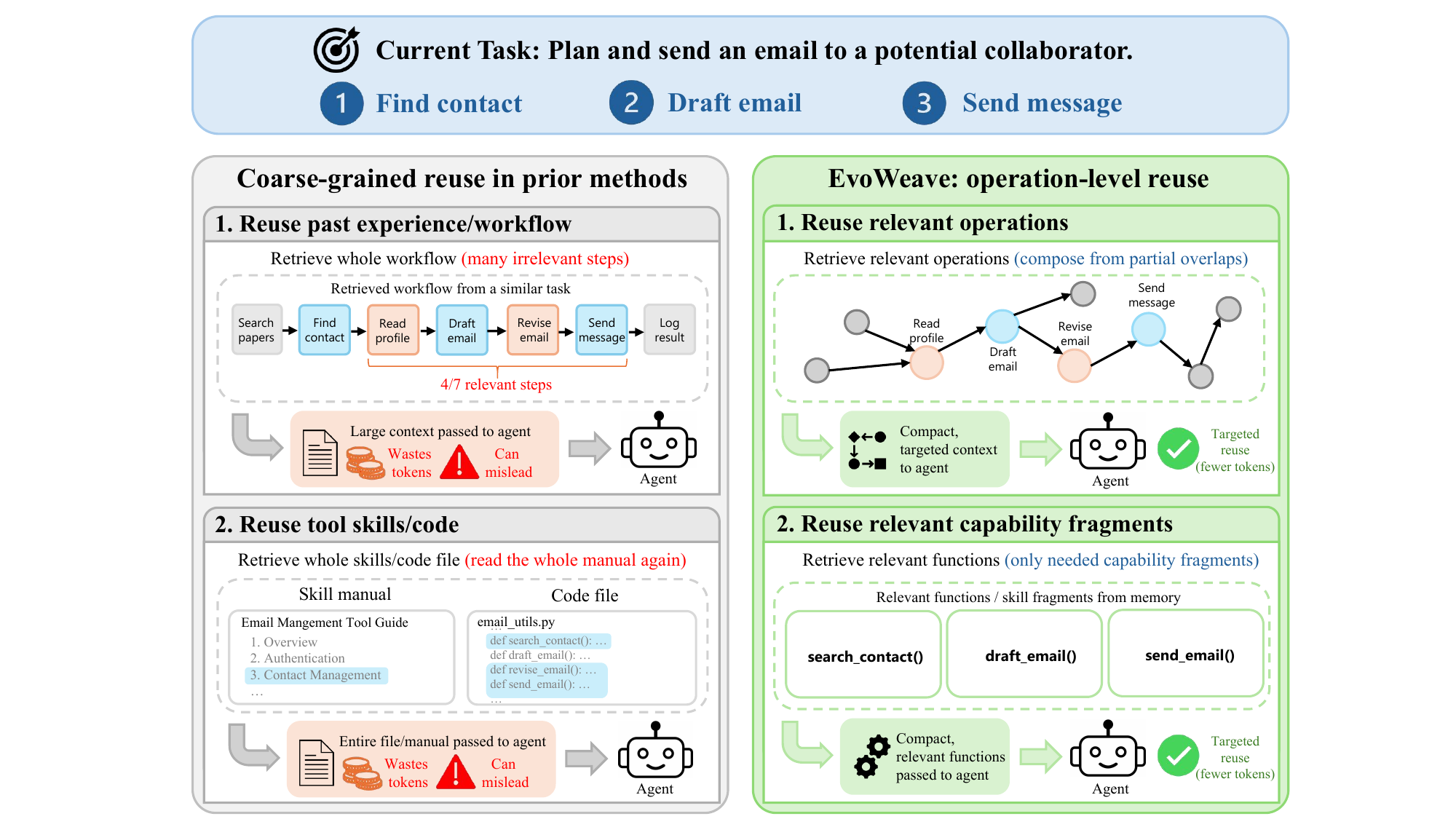}
\caption{Motivation for reusing individual operations and capability fragments. The illustrative comparison contrasts retrieving an entire historical workflow or capability file with selecting relevant operations and functions for a new task. Selective reuse aims to reduce irrelevant context while supporting composition across partially overlapping experiences. The figure is conceptual; its step counts and anticipated benefits are not experimental measurements or claims about every prior method.}
\label{fig:motivation}
\end{figure}

One part of that choice is granularity: the unit at which experience is stored and retrieved, which decides how much of a past task is carried into a new one. Figure~\ref{fig:motivation} illustrates this choice. A new request to contact a collaborator may require finding contact information, drafting an email, and sending a message. A historical workflow containing these operations, however, may also include unrelated research, profile inspection, or record keeping, so supplying that entire workflow couples useful experience to unnecessary context. We therefore propose retrieving the relevant operations separately, which allows partial reuse across historical tasks. We use three terms consistently: an \emph{operation} is one abstract step of a task, such as sending a message; an \emph{action} is the concrete environment call that realizes an operation during execution; and a \emph{capability} is the external resource that such an action needs, such as an API definition or a library function. The same granularity principle then applies to capabilities: an operation can reference only the API definition or function required by its action, rather than an entire manual or code library. Longer context does not guarantee that a model uses the relevant parts of it \citep{liu2024lost}, so we treat narrower retrieval as a hypothesis to be measured rather than a benefit to be assumed.

Fine grained reuse alone does not solve composition. By composition we mean assembling several separately retrieved operations into one executable sequence for the current task, such that every operation finds the inputs it requires either in the environment or in the outputs of the operations placed before it. Two individually relevant operations may disagree about the intermediate object they consume, or require an unobserved prerequisite. A path to a downstream operation does not establish that every input is available. Prior work has already explored reusable programs \citep{wang2023voyager,yuan2024craft}, linked memories \citep{xu2026mem}, and modular procedural memory \citep{han2025legomem}. What these representations do not retain is, for each reusable unit, the input conditions and the directed transitions that determine whether the unit can be executed after the units retrieved alongside it. The question we address is therefore more specific: how can operation level experience preserve its conditions and historical connections while remaining selectable for a new task?

We propose \method{}, which represents an operation as a directed transition between abstract states, annotated with its contract, evidence, and capability references. A current task plan specifies which steps directly depend on one another. These dependencies guide retrieval of historical operations and search for connecting fragments. The execution agent receives the assembled operation fragments as context, together with an explicit list of the plan dependencies that retrieval could not connect, and it can query the experience graph again during execution. Reflection retains concrete observations in task history while proposing abstract, versioned updates to shared experience.

Our contributions are threefold: (i) an operation level representation that separates reusable transitions from concrete execution evidence; (ii) a plan guided retrieval and connection procedure that retains input requirements, candidate consistency, and search failures; and (iii) an empirical analysis spanning three backbones and five evaluation settings, with ablations that distinguish operation decomposition, directed connections, and active retrieval. We report heterogeneous gains and resource tradeoffs rather than assuming that a graph representation improves every task.

\section{Related Work}
\paragraph{Experiential and procedural memory.}
ExpeL distills experiential knowledge \citep{zhao2024expel}, Reflexion uses verbal feedback \citep{shinn2023reflexion}, and ReasoningBank organizes reasoning memories from previous interactions \citep{ouyang2026reasoningbank}. CLIN maintains textual causal abstractions across trials, while Buffer of Thoughts retrieves reusable reasoning templates \citep{majumder2023clin,yang2024buffer}. Synapse retrieves trajectories with abstracted states, Agent Workflow Memory induces workflows, and LEGOMem studies modular procedural memory \citep{zheng2024synapse,wang2024agent,han2025legomem}. These approaches show that neither experience reuse nor decomposition originates with our work. \method{} focuses on operations as state transition edges and uses direct plan dependencies to retrieve and connect historical fragments. Our ablations examine these connections within the reported configurations.

\paragraph{Structured memory and retrieval.}
MemoryBank and Generative Agents connect persistent memories to later interactions and behavior \citep{zhong2024memorybank,park2023generative}. Mem0 and A-MEM investigate scalable storage and memory organization, while MemGPT manages context levels \citep{chhikara2025mem0,xu2026mem,packer2023memgpt}. Retrieval augmented generation provides external evidence; Self-RAG further learns when to retrieve and how to critique generations \citep{lewis2020retrieval,asai2024self}. GraphRAG and HippoRAG organize retrieval through graphs \citep{edge2024local,gutierrez2024hipporag}. Closer to interactive agents, AriGraph combines semantic and episodic memory in a world model, and G-Memory traces collaborative experience through a graph hierarchy \citep{anokhin2024arigraph,zhang2026g}. \method{} instead centers retrieval on operation contracts and historical transitions. Reachability supplies a candidate fragment, not a certificate that a live action is feasible.

\paragraph{Reusable capabilities and planning.}
Voyager, LATM, CRAFT, and Code as Policies explore reusable skills, tool creation, or executable programs \citep{wang2023voyager,cai2024large,yuan2024craft,liang2023code}. LearnAct improves action functions using execution feedback, while Toolformer studies learning tool use \citep{zhao2024empowering,schick2023toolformer}. Skill Retrieval Augmentation distinguishes retrieving a skill, incorporating it, and completing the task \citep{su2026skill}, supporting our separation of memory references from execution evidence. LLMCompiler and SwiftSage address planning and execution, and LATS searches action trajectories using environmental feedback \citep{kim2023llm,lin2023swiftsage,zhou2023language}. Our bounded search connects stored operations rather than simulating future actions, and it requires no update to backbone parameters. AgentBoard's analysis of progress beyond final success further motivates examining category behavior and reuse alongside aggregate scores \citep{ma2024agentboard}.

\section{Method: EvoWeave}
\label{sec:method}
\begin{figure}[t]
\centering
\includegraphics[width=\linewidth]{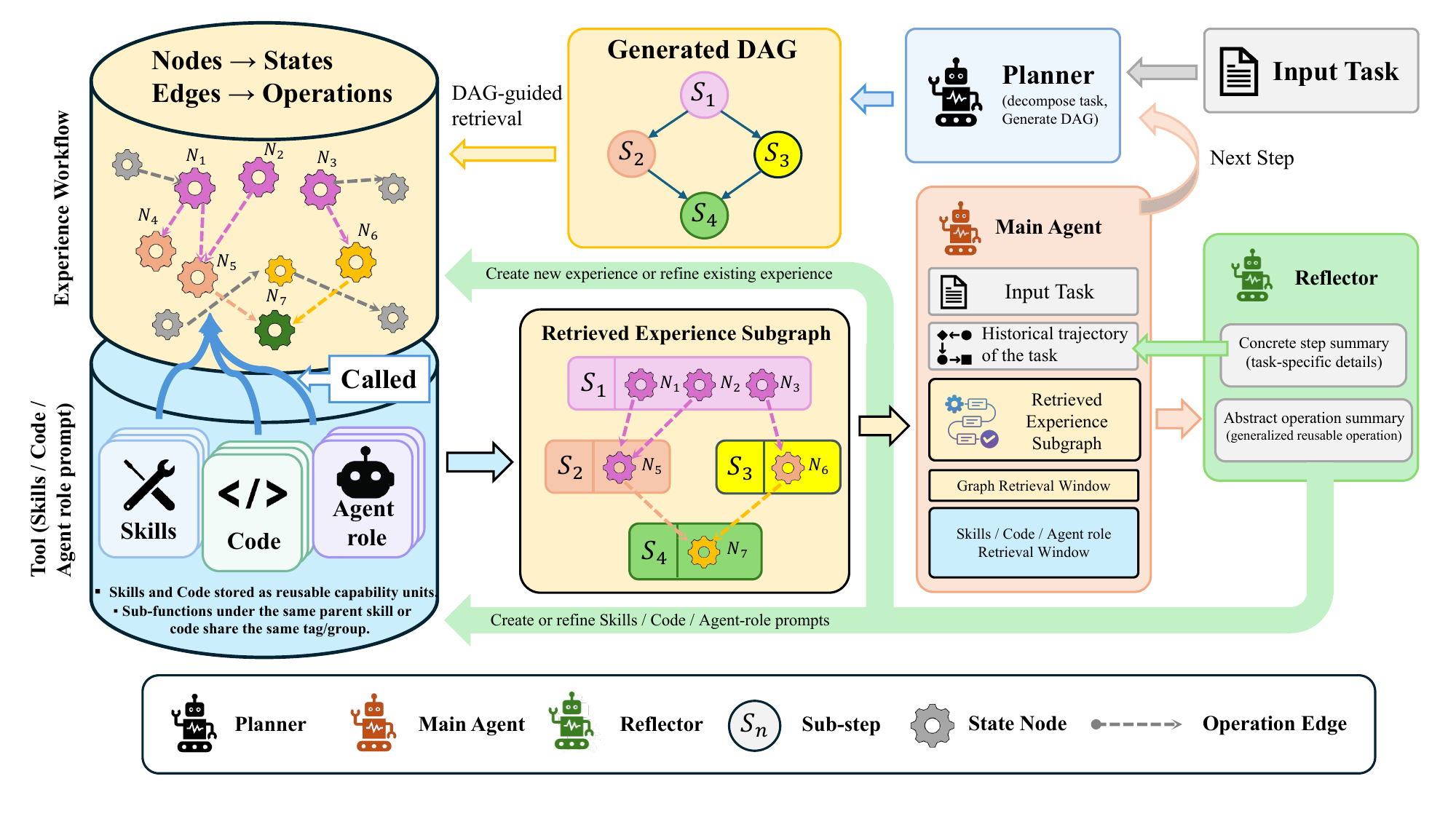}
\caption{Overview of \method{}. A planner generates a task DAG that guides retrieval from an experience graph, whose nodes represent states and edges represent operations. The main agent uses retrieved experience and relevant capabilities, while the reflector separates concrete step summaries from abstract operation summaries. Shared memory updates follow the declared construction or evaluation protocol. Capability and role refinement in the diagram describe the broader architecture; evaluated implementation boundaries are detailed in Appendix~\ref{app:implementation}.}
\label{fig:method}
\end{figure}

Figure~\ref{fig:method} shows the flow from task planning to experience retrieval, capability access, execution, and reflection.
\subsection{Experience as Operations with Explicit Contracts}
Given a task $q$, an environment, and a concrete interaction history $H_t$, the agent must produce actions within an interaction budget. We distinguish its current plan $P_t$, which is a directed acyclic graph (DAG), from the historical experience graph $G=(V,E)$, which can contain cycles and multiple edges between states. A reusable operation is
\begin{equation}
e=(u,v,d,\mathcal I,\mathcal O,\mathcal C,\mathcal R,\pi),
\end{equation}
where $u,v$ are abstract source and target states, $d$ is an operation description, $\mathcal I$ and $\mathcal O$ specify inputs and outputs, $\mathcal C$ records conditions and effects, $\mathcal R$ contains capability references, and $\pi$ records provenance and versions. Task specific entities and values become placeholders in shared experience; their actual bindings remain in $H_t$. An operation with several required inputs retains their conjunction, including when a retrieved path supplies only one of them.

\subsection{Retrieval and Connections Guided by the Plan}
The planner uses $q$ and $H_t$ to specify remaining steps $S_t$ and their direct dependencies $D_t$, without first conditioning the plan on available memory. For each step $s_i$, operation retrieval uses description embeddings $f$:
\begin{equation}
C_i=\operatorname{TopK}_{k=3}\!\left\{e\in E:\cos\big(f(s_i),f(d_e)\big)\geq\tau_{\mathrm{retrieve}}\right\}.
\end{equation}
Filtering precedes ranking, and an empty candidate set is valid. Candidates retain edge identities, endpoints, and contracts; operations sharing a source state remain distinct.

For each direct dependency $(i,j)\in D_t$, consider a candidate pair $a\in C_i$, $b\in C_j$. A connection searches from the target state $v_a$ to the source state $u_b$ and retains both matched operations:
\begin{equation}
\Gamma_{ij}=a\ \circ\ \operatorname{BFS}_{G}(v_a,u_b)\ \circ\ b.
\end{equation}
Breadth first search minimizes the number of connecting operation edges under fixed search limits. At most nine candidate pairs are examined per plan dependency. Ties use candidate similarity and stable identifiers. We do not add connection requests for transitive plan dependencies. An unsuccessful search reports either a gap or a budget truncation, instead of inventing an edge.

Connections sharing a plan step must use a consistent candidate to form one chain. If local choices cannot be reconciled, we retain separate fragments and expose a candidate conflict. This conservative procedure is not a global assignment of minimum total cost. Similarly, a graph path does not establish that all inputs are bound, entities are consistent, or concurrent writes are safe. The main agent checks these conditions against actual observations before acting.

\subsection{Execution and Capability Access}
The main agent receives the plan, retrieved operation cards, connecting fragments, and missing conditions. Before opening capability bodies, it can inspect an operation, retrieve neighbors, search descriptions, or request a path. Cards expose identifiers, contracts, provenance, and versions; subsequent requests load the relevant API or function body. Official API definitions remain immutable, with learned usage information stored separately. New code capabilities require validation and tests. The agent generates current task actions using these references and can reject a retrieved suggestion or request replanning. Events recording memory references are distinct from evidence that a stored program was directly executed.

\subsection{Reflection and Controlled Memory Updates}
Following execution, reflection produces two distinct artifacts. Concrete history records actual inputs, outputs, errors, and evidence references. Abstract experience records reusable descriptions, placeholders, conditions, and capability references. Every update branch performs this abstraction, including additions to existing operations. A separate similarity threshold $\tau_{\mathrm{merge}}$ selects candidates for comparison; similarity alone does not authorize a merge. Contract checks distinguish compatible evidence additions, variants, and new operations. For example, requesting a payment and transferring money must remain distinct despite lexical similarity.

Failures retain the observed error and any evidenced recovery, without promoting expected outputs into successful outcomes. Embedding failures are distinguished from legitimate retrieval without a match. Memory proposals undergo structural, capability, and contract validation before publication. Having three persistent roles (planner, main, and reflector) does not imply exactly three model calls per round: active retrieval and merge decisions may introduce additional calls. Appendix~\ref{app:implementation} provides pseudocode and default parameters for the reference implementation.

\section{Experiments}
\label{sec:experiments}
\subsection{Benchmarks, Backbones, and Reporting Protocol}
We analyze five settings from four benchmarks: AppWorld Test-Normal and Test-Challenge \citep{trivedi2024appworld}, WebArena \citep{zhou2024webarena}, ALFWorld \citep{shridhar2020alfworld}, and ScienceWorld \citep{wang2022scienceworld}. AppWorld reports task goal completion (TGC) and scenario goal completion (SGC); WebArena reports success rate (SR). ALFWorld and ScienceWorld are reported with their own benchmark specific scores, whose scoring semantics need not be identical. Higher values are better for all seven columns. We average matching metrics equally over backbones and never combine heterogeneous metrics into one score. Differences are computed before rounding; display values are rounded afterward.

The backbones are DeepSeek-V3.2 \citep{liu2025deepseek}, GPT-4.1-mini, and Qwen3.8-27B. Qwen is served locally on an institutional HPC cluster; DeepSeek-V3.2 and GPT-4.1-mini are accessed through API calls. The analysis uses aggregate results for these configurations. Token budgets are not equalized by construction across methods; Appendix~\ref{app:efficiency} reports the tokens each one actually consumes.

We compare No-Memory and Flat-Memory with AWM \citep{wang2024agent}, ReasoningBank \citep{ouyang2026reasoningbank}, Mem0 \citep{chhikara2025mem0}, ExpeL \citep{zhao2024expel}, and A-MEM \citep{xu2026mem}. These comparisons concern the evaluated implementations and do not establish the maximum performance attainable by each method under other adaptations or budgets.

\paragraph{Memory construction protocol.}
No memory is built from training or development data. Every run starts from an empty memory and accumulates experience online during evaluation: test tasks are executed one at a time in the order in which the benchmark releases them, and experience written after a task becomes available to all later tasks in the same run. Each of the 15 combinations of backbone and evaluation setting is an independent run that starts empty, so no experience crosses backbones or benchmarks. Reflection records what the agent observes within an episode, namely API return values, environment errors, and observations, together with an episode level success judgement that a language model derives from the trajectory itself. Methods that need a success signal, such as ExpeL and AWM, take it from this same module, so no method receives a more reliable signal than another. The official task level score is computed after an episode terminates, is used only for reporting, and never reaches the memory of any method. Every memory based baseline runs under exactly this protocol, from the same empty start and the same task order, so the comparison isolates how experience is organized rather than how much data each method was allowed to see. The setting is therefore continual learning over an evaluation stream, applied identically to every method; Appendix~\ref{app:protocol} records the protocol in full.

Main results are reported as mean $\pm$ standard deviation (SD). We report descriptive differences without significance claims. Appendix~\ref{app:verification} details the evaluation scope, aggregation rules, and comparability of benchmark and category summaries.

\subsection{Main Results}
\begin{table}[t]
\caption{Main results reported as mean $\pm$ standard deviation (SD), with the $\pm$ SD terms shown as subscripts. All entries retain the original score scale from 0 to 100. Within each backbone block, the highest mean in a column is \colorbox{gray!25}{shaded} and the second highest is \underline{underlined}; tied values receive the same mark.}
\label{tab:main}
\centering
\begingroup\footnotesize\setlength{\tabcolsep}{1.8pt}
\begin{tabular}{lrrrrrrr}
\toprule
& \multicolumn{2}{c}{AppWorld Normal} & \multicolumn{2}{c}{AppWorld Challenge} & WebArena & ALFWorld & ScienceWorld \\
\cmidrule(lr){2-3}\cmidrule(lr){4-5}
Method & TGC & SGC & TGC & SGC & SR & Score & Score \\
\midrule
\multicolumn{8}{l}{\textit{DeepSeek-V3.2}} \\
No-Memory & $47.02_{\pm 2.75}$ & $30.36_{\pm 3.30}$ & $23.02_{\pm 1.97}$ & $9.35_{\pm 1.83}$ & $42.00_{\pm 1.36}$ & $69.40_{\pm 2.41}$ & $64.70_{\pm 2.52}$ \\
Flat-Memory & $49.05_{\pm 3.03}$ & $32.14_{\pm 3.68}$ & $23.02_{\pm 2.16}$ & $9.35_{\pm 2.01}$ & $43.00_{\pm 1.49}$ & $71.94_{\pm 2.59}$ & $67.20_{\pm 2.77}$ \\
AWM & \underline{$52.02_{\pm 2.92}$} & \underline{$35.00_{\pm 3.62}$} & $24.99_{\pm 2.12}$ & $10.50_{\pm 1.99}$ & \underline{$46.01_{\pm 1.44}$} & $73.88_{\pm 2.47}$ & $68.70_{\pm 2.67}$ \\
ReasoningBank & $52.01_{\pm 3.99}$ & \underline{$35.00_{\pm 4.95}$} & \underline{$26.00_{\pm 2.92}$} & \cellcolor{gray!25}$11.88_{\pm 2.76}$ & $43.00_{\pm 1.97}$ & $71.34_{\pm 3.46}$ & $67.70_{\pm 3.66}$ \\
Mem0 & $47.50_{\pm 3.08}$ & $30.71_{\pm 3.71}$ & $22.49_{\pm 2.19}$ & $9.06_{\pm 2.04}$ & $42.51_{\pm 1.52}$ & $69.40_{\pm 2.69}$ & $65.20_{\pm 2.82}$ \\
ExpeL & $50.95_{\pm 2.95}$ & $33.93_{\pm 3.63}$ & $24.99_{\pm 2.14}$ & $10.50_{\pm 2.01}$ & $45.00_{\pm 1.46}$ & \cellcolor{gray!25}$76.12_{\pm 2.45}$ & \underline{$74.10_{\pm 2.70}$} \\
A-MEM & $47.98_{\pm 3.25}$ & $31.07_{\pm 3.92}$ & $23.02_{\pm 2.32}$ & $9.35_{\pm 2.16}$ & $43.00_{\pm 1.60}$ & $70.90_{\pm 2.81}$ & $66.70_{\pm 2.98}$ \\
EvoWeave & \cellcolor{gray!25}$54.05_{\pm 2.88}$ & \cellcolor{gray!25}$37.14_{\pm 3.63}$ & \cellcolor{gray!25}$27.00_{\pm 2.13}$ & \underline{$11.65_{\pm 2.02}$} & \cellcolor{gray!25}$49.00_{\pm 1.43}$ & \underline{$75.88_{\pm 2.45}$} & \cellcolor{gray!25}$76.20_{\pm 2.65}$ \\
\midrule
\multicolumn{8}{l}{\textit{GPT-4.1-mini}} \\
No-Memory & $35.71_{\pm 3.18}$ & $20.71_{\pm 3.53}$ & $16.98_{\pm 2.22}$ & $6.33_{\pm 2.03}$ & $30.69_{\pm 1.58}$ & $45.97_{\pm 3.07}$ & $40.00_{\pm 3.03}$ \\
Flat-Memory & $36.67_{\pm 3.52}$ & $21.43_{\pm 3.93}$ & $16.02_{\pm 2.42}$ & $5.90_{\pm 2.21}$ & $30.69_{\pm 1.74}$ & $48.96_{\pm 3.37}$ & $41.00_{\pm 3.33}$ \\
AWM & $38.69_{\pm 3.42}$ & $22.86_{\pm 3.86}$ & $17.99_{\pm 2.38}$ & $6.76_{\pm 2.18}$ & $33.69_{\pm 1.69}$ & $51.04_{\pm 3.27}$ & $44.00_{\pm 3.21}$ \\
ReasoningBank & \underline{$39.71_{\pm 3.42}$} & \underline{$24.64_{\pm 3.92}$} & \underline{$20.00_{\pm 2.41}$} & \cellcolor{gray!25}$7.77_{\pm 2.22}$ & \underline{$35.69_{\pm 1.69}$} & $51.94_{\pm 3.23}$ & $46.00_{\pm 3.18}$ \\
Mem0 & $35.71_{\pm 3.57}$ & $20.71_{\pm 3.95}$ & $16.02_{\pm 2.46}$ & $5.90_{\pm 2.25}$ & $30.69_{\pm 1.77}$ & $45.07_{\pm 3.42}$ & $40.00_{\pm 3.39}$ \\
ExpeL & $38.69_{\pm 3.46}$ & $22.86_{\pm 3.90}$ & $17.99_{\pm 2.41}$ & $6.76_{\pm 2.20}$ & $32.71_{\pm 1.70}$ & \underline{$54.03_{\pm 3.29}$} & \underline{$48.00_{\pm 3.24}$} \\
A-MEM & $36.67_{\pm 3.78}$ & $21.43_{\pm 4.21}$ & $16.98_{\pm 2.62}$ & $6.33_{\pm 2.40}$ & $31.70_{\pm 1.87}$ & $48.06_{\pm 3.62}$ & $42.00_{\pm 3.57}$ \\
EvoWeave & \cellcolor{gray!25}$40.60_{\pm 4.59}$ & \cellcolor{gray!25}$26.79_{\pm 5.03}$ & \cellcolor{gray!25}$21.01_{\pm 3.15}$ & \underline{$7.47_{\pm 2.87}$} & \cellcolor{gray!25}$36.70_{\pm 2.29}$ & \cellcolor{gray!25}$57.97_{\pm 4.47}$ & \cellcolor{gray!25}$51.00_{\pm 4.39}$ \\
\midrule
\multicolumn{8}{l}{\textit{Qwen3.8-27B}} \\
No-Memory & $52.02_{\pm 3.03}$ & $35.00_{\pm 3.75}$ & $29.02_{\pm 2.26}$ & $12.95_{\pm 2.17}$ & $43.99_{\pm 1.50}$ & $71.94_{\pm 2.60}$ & $66.10_{\pm 2.77}$ \\
Flat-Memory & $54.05_{\pm 3.32}$ & $37.14_{\pm 4.18}$ & $30.02_{\pm 2.50}$ & $13.53_{\pm 2.41}$ & $45.00_{\pm 1.64}$ & $74.03_{\pm 2.80}$ & $68.00_{\pm 3.05}$ \\
AWM & $57.02_{\pm 3.18}$ & $40.36_{\pm 4.09}$ & $30.98_{\pm 2.42}$ & $14.24_{\pm 2.35}$ & $48.00_{\pm 1.59}$ & $75.97_{\pm 2.67}$ & $70.50_{\pm 2.94}$ \\
ReasoningBank & \underline{$57.98_{\pm 3.15}$} & \underline{$41.43_{\pm 4.06}$} & \underline{$33.00_{\pm 2.42}$} & \underline{$15.54_{\pm 2.38}$} & \cellcolor{gray!25}$50.00_{\pm 1.58}$ & $77.01_{\pm 2.61}$ & $72.20_{\pm 2.91}$ \\
Mem0 & $52.98_{\pm 3.39}$ & $36.07_{\pm 4.23}$ & $29.02_{\pm 2.53}$ & $12.95_{\pm 2.43}$ & $45.00_{\pm 1.67}$ & $71.94_{\pm 2.90}$ & $67.00_{\pm 3.11}$ \\
ExpeL & $55.95_{\pm 3.22}$ & $39.29_{\pm 4.11}$ & $30.98_{\pm 2.44}$ & $14.24_{\pm 2.37}$ & $47.00_{\pm 1.60}$ & \underline{$78.96_{\pm 2.61}$} & \cellcolor{gray!25}$74.10_{\pm 2.97}$ \\
A-MEM & $54.05_{\pm 3.57}$ & $37.14_{\pm 4.48}$ & $30.02_{\pm 2.68}$ & $13.53_{\pm 2.58}$ & $46.01_{\pm 1.77}$ & $74.03_{\pm 3.01}$ & $69.00_{\pm 3.27}$ \\
EvoWeave & \cellcolor{gray!25}$61.07_{\pm 4.36}$ & \cellcolor{gray!25}$43.92_{\pm 5.59}$ & \cellcolor{gray!25}$35.97_{\pm 3.33}$ & \cellcolor{gray!25}$17.82_{\pm 3.25}$ & \underline{$49.00_{\pm 2.17}$} & \cellcolor{gray!25}$79.03_{\pm 3.72}$ & \underline{$73.70_{\pm 4.02}$} \\
\bottomrule
\end{tabular}

\endgroup
\end{table}

Table~\ref{tab:main} shows that \method{} has the highest reported mean in \MainWins{} of the 21 combinations of backbone and metric. Across backbones, Normal TGC increases from \NoNormal{} to \MainNormal{} and Challenge TGC from \NoChallenge{} to \MainChallenge{} relative to No-Memory. In both cases the strongest baseline macro mean comes from ReasoningBank, over which \method{} gains \BestNormalGain{} and \BestChallengeGain{} points, respectively. Averages across backbones for WebArena SR, ALFWorld Score, and ScienceWorld Score reach \MainWeb{}, \MainAlf{}, and \MainScience{}, respectively, compared with \NoWeb{}, \NoAlf{}, and \NoScience{} without memory.

The gains are not uniform. ReasoningBank exceeds \method{} on DeepSeek and GPT Challenge SGC and on Qwen WebArena SR. ExpeL has the higher mean on DeepSeek ALFWorld and Qwen ScienceWorld. Some differences are small relative to the reported standard deviations. These exceptions matter: matching and connecting historical operations does not guarantee that the selected experience fits the current task or improves every backbone.

\subsection{Category Performance}
\begin{figure}[t]
\centering
\includegraphics[width=\linewidth]{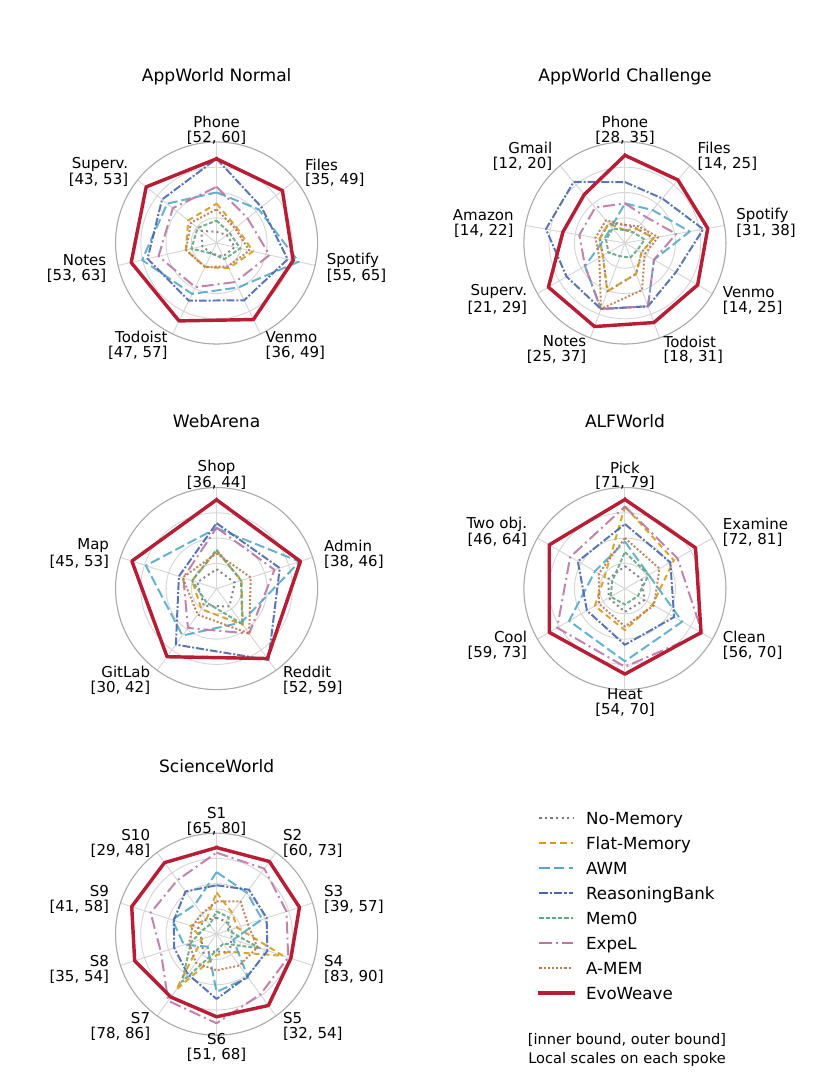}
\caption{Category performance averaged equally over three backbones. Each spoke uses a local linear range determined by all eight methods; brackets give its inner and outer bounds in original score units. Compare methods on the same spoke: distances across spokes, panels, and polygon areas are not directly comparable. Labels and the scaling rule are defined in Appendix~\ref{app:radar}. These category aggregates are distinct from Table~\ref{tab:main}; see Appendix~\ref{app:verification}.}
\label{fig:radars}
\end{figure}
Figure~\ref{fig:radars} separates category behavior from benchmark averages. For \method{}, AppWorld Challenge ranges from \CatChallengeMin{} on Gmail to \CatChallengeMax{} on Spotify, while ALFWorld ranges from \CatAlfMin{} for Use Two Objects to \CatAlfMax{} for Examine in Light. ScienceWorld varies from \CatScienceMin{} for Mendelian genetics to \CatScienceMax{} for Living and Non-living Things. Thus, a strong aggregate does not imply uniform performance. The local axis ranges expose differences among methods without altering the original scores. All 15 panels for individual backbones appear in Appendix~\ref{app:radar}, preserving reversals and the original Supervisor category. WebArena category means weighted by task counts sit slightly above the main table because the five site categories cover 764 of the 812 tasks and exclude the 48 cross site tasks; Appendix~\ref{app:verification} quantifies that difference for each method.

\subsection{Ablation: Decomposition, Connections, and Active Retrieval}
\begin{table}[t]
\centering
\caption{AppWorld ablation, averaged equally over backbones. Overall TGC uses weights based on task counts of 168 Normal and 417 Challenge tasks within each backbone. These configurations are specific to AppWorld; only point estimates are shown. Configurations are ordered so that the full system appears last, and the highest value in each column is shown in bold.}
\label{tab:ablation}
\small\begin{tabular}{lrrr}
\toprule
Configuration & Normal TGC & Challenge TGC & Overall TGC \\
\midrule
A: No memory & 44.92 & 23.01 & 29.30 \\
B: Whole workflow RAG & 46.59 & 23.02 & 29.79 \\
C: Operation cards & 47.93 & 24.72 & 31.39 \\
D: C + directed connections & 50.74 & 26.51 & 33.47 \\
E: C + active retrieval & 49.90 & 25.94 & 32.82 \\
F: EvoWeave (full) & \textbf{51.91} & \textbf{27.99} & \textbf{34.86} \\
\bottomrule
\end{tabular}

\end{table}
Table~\ref{tab:ablation} reports six configurations. A removes memory and B retrieves whole workflows, so they coincide with the No-Memory and Flat-Memory rows of Table~\ref{tab:main}; C retrieves independent operation cards; D adds directed connections to C; E adds active retrieval to C without historical connections; and F is the full system, which adds active graph retrieval to D. Appendix~\ref{app:ablation} reports every backbone separately.

Operation cards improve Overall TGC from \AblB{} for whole workflow retrieval to \AblC{}. Adding directed connections further raises it to \AblD{}, a difference of \AblCDGain{} points. Active retrieval without connections reaches \AblE{}, improving on C by \AblCEGain{} points. The full system reaches \AblF{}, another \AblDFGain{} points above D and \AblFEGap{} points above E. These comparisons support the hypothesis that relational context contributes beyond decomposition and additional lookup opportunities. The ordering of D and E is not reproduced by every backbone: E slightly exceeds D on DeepSeek, while GPT and Qwen keep D above E (Appendix~\ref{app:ablation}). All six configurations share the memory construction protocol of Appendix~\ref{app:protocol}, the same candidate pool, and the same capability library, so they differ only in how retrieved experience is connected and queried.

\subsection{Memory Structure and Operation Reuse}
\begin{table}[t]
\centering
\caption{Mean memory size over the five settings. Counts describe the reported graph snapshots, not an inferred average topology. Capability links connect the corresponding node type to operations. Counts of role nodes do not establish that learning of role prompts was evaluated.}
\label{tab:graph}
\small\begin{tabular}{lrrrr}
\toprule
Experience layer & States & Operations & & \\
Mean count & 27.0 & 157.6 & & \\
\midrule
Capability layer & Code & Skills & Roles & \\
Mean nodes & 185.2 & 82.2 & 6.8 & \\
Mean links to operations & 157.6 & 362.0 & 99.4 & \\
\bottomrule
\end{tabular}

\end{table}
Table~\ref{tab:graph} summarizes graph size. An average snapshot contains \GraphStates{} state nodes and \GraphOperations{} operation edges; capability references extend across code, skills, and role nodes. These values describe the separation of transitions from the capabilities they reference, but do not measure successful composition or graph quality. The counts for individual settings vary substantially: AppWorld Challenge has 431 operations, compared with 14 in ALFWorld (Appendix~\ref{app:graph}). These arithmetic summaries characterize size and composition without defining a shared adjacency structure.

\begin{figure}[t]
\centering
\includegraphics[width=.64\linewidth]{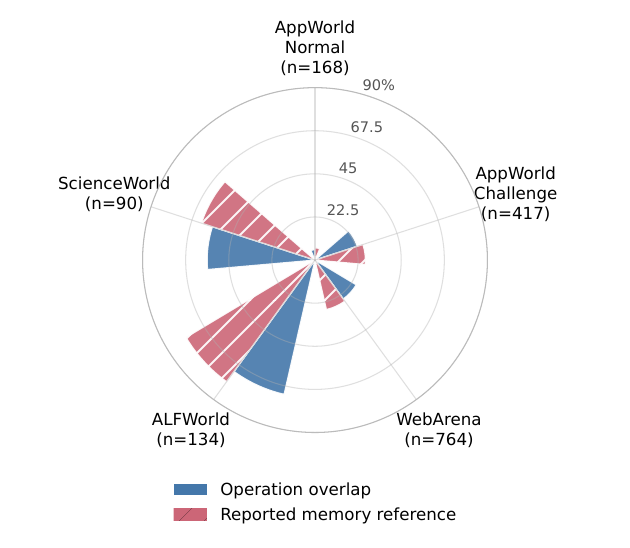}
\caption{Mean operation overlap and reported memory referencing across task types. Each type receives equal weight; overlap comparisons are restricted to tasks of the same type. Labels show type counts, and bars show percentages on a scale from 0 to 100. Benchmark specific overlap proxies and reference rates are distinct measures (Appendix~\ref{app:overlap}).}
\label{fig:overlap}
\end{figure}

Figure~\ref{fig:overlap} summarizes recurring operations within task types. The equal type means of overlap/reference (\%) are \OverlapNormal{}/\ReferenceNormal{} for AppWorld Normal, \OverlapChallenge{}/\ReferenceChallenge{} for Challenge, \OverlapWeb{}/\ReferenceWeb{} for WebArena, \OverlapAlf{}/\ReferenceAlf{} for ALFWorld, and \OverlapScience{}/\ReferenceScience{} for ScienceWorld. Cross type comparisons are excluded from the overlap averages. These substantial within type overlaps motivate operation level abstraction: related tasks can reuse shared steps even when complete workflows differ. Appendix~\ref{app:overlap} defines the averages and provides all type specific values.

\subsection{Context and Runtime Tradeoffs}
The efficiency results show \InjectOurs{} injected memory tokens per question for \method{}, versus \InjectFlat{} for Flat-Memory, \InjectAWM{} for AWM, and \InjectRB{} for ReasoningBank: reductions of \SaveFlat{}\%, \SaveAWM{}\%, and \SaveRB{}\%, respectively. These figures support selective context delivery, but the total input token measure falls by only \EnterSaving{}\% relative to No-Memory. \method{} has the fewest rounds (\RoundsOurs{}) and the highest seconds per round (\SecondsOurs{}), alongside lower KV reuse (\KVOurs{}\%). Less injected memory therefore does not by itself demonstrate lower latency or cost. Appendix~\ref{app:efficiency} gives all efficiency metrics and explains the boundaries of token and runtime comparisons.

\section{Conclusion}
\method{} organizes reusable experience as operation edges with explicit contracts and connects retrieved fragments according to a current task dependency plan. This design preserves local reuse opportunities while exposing missing conditions and inconsistent connections. Results across three backbones and five settings are broadly favorable, and AppWorld ablations associate both operation decomposition and directed connections with improved task completion. The framework also highlights that smaller memory context can incur additional retrieval and reasoning overhead. Further work should examine generalization to unseen operation compositions and the effect of retrieval overhead on end to end performance.

\label{maintextend}
\section*{Ethics Statement}
This work follows the ICLR Code of Ethics. No human participants or animal experiments were involved in this study. All benchmarks used in this work, namely AppWorld, WebArena, ALFWorld, and ScienceWorld, are publicly released research benchmarks that run in simulated environments, and we used them in accordance with their respective licenses and usage policies. Throughout the research process, we made efforts to minimize potential biases and prevent discriminatory outcomes. No personally identifiable information was collected or utilized, and no experiments were performed that would pose potential privacy or security risks. We remain committed to ensuring transparency, ethical conduct, and research integrity throughout the study.

\section*{Reproducibility Statement}
We have taken the following steps to make this work reproducible. The operation representation, the plan guided retrieval and connection procedure, and the reflection and memory update rules are defined in Section~\ref{sec:method}; Appendix~\ref{app:implementation} gives the corresponding pseudocode (Algorithm~\ref{alg:retrieval}), the complexity bound on the number of bounded searches, and the complete default parameters of the reference implementation (Table~\ref{tab:defaults}), including retrieval and merge thresholds, candidate counts, search depth and expansion limits, and the budget for visible operation cards. All four benchmarks are publicly available, and we use their released task splits without modification; the benchmarks, backbones, baseline methods, metrics, and aggregation rules are enumerated in Section~\ref{sec:experiments} and summarized in Table~\ref{tab:metadata}. Appendix~\ref{app:verification} states the aggregation formulas, the task counts used for weighted means, and the exact sense in which category, graph, overlap, and efficiency summaries are computed over populations that differ from the main benchmark means. Appendices~\ref{app:ablation} to~\ref{app:overlap} report the per backbone tables, per setting graph statistics, token and runtime metrics, and category level overlap definitions underlying every figure in the main text.
Two limitations bound exact reproduction and we state them explicitly. First, decoding is stochastic and two of the three backbones are served through provider APIs whose deployed checkpoints we do not control, so individual runs will not match digit for digit; we therefore report means with standard deviations and make only descriptive comparisons, as explained in Appendix~\ref{app:verification}. Second, the baselines are re-implementations evaluated under our harness rather than numbers copied from their original papers. They all run under the common memory construction protocol of Appendix~\ref{app:protocol}, so the comparison holds the memory source, the task order, the success judgement module, and the per run isolation fixed. Token budgets are not equalized by construction; Appendix~\ref{app:efficiency} reports the injected and total input tokens of every method, and \method{} has the smallest counts of both, so its results are not obtained from a larger context.

\section*{AI Use Statement}
Large Language Models (LLMs) were utilized to support the writing and language refinement of this manuscript. In particular, the LLMs were employed to improve the clarity, readability, and overall linguistic quality of the text across different sections of the paper. Their assistance included rephrasing sentences, correcting grammatical issues, and improving the coherence and flow of the manuscript.

It should be emphasized that the LLM was not involved in developing the research ideas, designing the methodology, or conducting the experiments. All research questions, concepts, methodological decisions, and analyses were independently developed and carried out by the authors. The role of the LLM was strictly limited to language-related assistance and did not extend to the scientific content, experimental analysis, or interpretation of the data.

The authors remain fully responsible for all content presented in the manuscript, including any text that was generated or subsequently refined with the assistance of an LLM. All LLM-assisted text was reviewed by the authors to ensure compliance with relevant ethical standards and to avoid plagiarism or any form of scientific misconduct.

\bibliography{references}
\bibliographystyle{iclr2027_conference}
\clearpage
\appendix
\setcounter{topnumber}{4}
\setcounter{bottomnumber}{3}
\setcounter{totalnumber}{6}
\renewcommand{\topfraction}{0.92}
\renewcommand{\bottomfraction}{0.85}
\renewcommand{\textfraction}{0.07}
\renewcommand{\floatpagefraction}{0.72}
\section{Evaluation Scope and Statistical Reporting}
\label{app:verification}
This appendix describes the scope and aggregation of the experimental results. The analysis combines benchmark summaries, category scores, ablations, graph statistics, and efficiency measurements. These summaries answer different questions and need not have identical task populations or aggregation rules. We therefore interpret each metric within its evaluation setting and distinguish descriptive comparisons from statistical significance.

\subsection{Evaluation Scope}
\begin{table}[htbp]
\centering\small
\begin{tabular}{p{.22\linewidth}p{.70\linewidth}}
\toprule
Item & Evaluation scope and reporting convention \\
\midrule
Benchmarks & AppWorld Test-Normal and Test-Challenge, WebArena, ALFWorld, and ScienceWorld. These give five settings from four benchmarks. \\
Backbones & DeepSeek-V3.2, GPT-4.1-mini, and Qwen3.8-27B. \\
Deployment & Qwen served locally on an institutional HPC cluster; DeepSeek-V3.2 and GPT-4.1-mini accessed through API calls. \\
Methods & No-Memory, Flat-Memory, AWM, ReasoningBank, Mem0, ExpeL, A-MEM, and EvoWeave. \\
Main metrics & AppWorld TGC and SGC, WebArena SR, and separate scores for ALFWorld and ScienceWorld, all on scales from 0 to 100. \\
Memory construction & Online over the evaluation stream. Every run starts from an empty memory, uses the released benchmark task order, and is independent across the 15 backbone and setting combinations. No training or development data is used, and episode level success is judged by a language model module shared by all methods rather than by the benchmark scorer. \\
Aggregation & Matching metrics are averaged equally over backbones. AppWorld Overall TGC weights Normal and Challenge by 168 and 417 tasks, respectively. \\
Uncertainty & Main results show mean $\pm$ standard deviation. Ablations show point estimates. Cross-method differences are descriptive. \\
Auxiliary analyses & Category performance, graph composition, token and runtime statistics, and operation overlap are analyzed separately from benchmark means. \\
\bottomrule
\end{tabular}
\caption{Evaluation scope and reporting conventions.}
\label{tab:metadata}
\end{table}

The AppWorld ablation combines Normal and Challenge using the two test counts in Table~\ref{tab:metadata}. This arithmetic relation does not by itself establish the identity of the evaluated tasks. Experience is acquired online across test tasks under the protocol of Section~\ref{sec:experiments}, which is applied identically to every memory based method; Appendix~\ref{app:protocol} records it in full.

\subsection{Benchmark and Category Aggregation}
Benchmark means and category summaries are treated as distinct measurements. For ScienceWorld, weighting category scores by their task counts reproduces the main means to numerical precision. This relation does not hold uniformly across all benchmarks.

For AppWorld, all 48 comparisons between main TGC and the corresponding Supervisor score agree to within 0.014 points, so the two views of the same runs are mutually consistent. The category analysis nevertheless keeps the Supervisor score as a distinct measure rather than treating it as interchangeable with main TGC, because the two are aggregated over different populations.

For WebArena, the main table reports success rate over the full benchmark of 812 tasks \citep{zhou2024webarena}. The category analysis covers the 764 tasks that belong to a single site, namely Shopping, Shopping Admin, GitLab, Map, and Reddit, whose counts appear in Figure~\ref{fig:overlap-web}. The remaining 48 tasks span more than one site and therefore belong to no single site category. Weighting the five category scores by their task counts consequently yields values \WebDiagMin{} to \WebDiagMax{} points above the main means: the category weighted mean omits the cross site tasks, on which every configuration scores lower than on single site tasks. For a fixed task split this difference is proportional to the gap between a method's single site and cross site success rates, and it is smallest for \method{} in all three backbones, at most \WebDiagOursMax{} points against at least \WebDiagBaseMin{} points for every baseline. Among the compared methods, \method{} therefore has the narrowest gap between tasks confined to one site and tasks that must be composed across sites. ALFWorld means weighted by category counts differ from the main table by approximately $\AlfDiagMin{}$ to $+\AlfDiagMax{}$ points. We therefore use category plots to describe variation within categories, without deriving additional benchmark means from them.

AppWorld category memberships in the overlap analysis sum to 183 for Normal and 795 for Challenge, exceeding their respective task totals. Memberships can overlap, so these category counts do not form a partition. Moreover, a global pairwise overlap measure includes tasks from different categories, whereas within category overlap excludes those pairs. The global Jaccard mean is therefore not generally equal to an average of within category Jaccard means, even when weighted by task counts. Figure~\ref{fig:overlap} specifically averages within type overlap with equal weight per type.

\subsection{Memory Construction and Leakage Controls}
\label{app:protocol}
Section~\ref{sec:experiments} states the protocol; this subsection records it in full. Experience is acquired online: a run visits the test tasks of one evaluation setting in sequence, and whatever the reflector writes after a task is available to the tasks that follow it in the same run. The setting is therefore continual learning over an evaluation stream, and it is the setting in which reusable experience is intended to be useful.

Three properties fix what the protocol can and cannot draw on. First, no task is attempted twice, and the agent receives no ground truth for the task it is solving. Several of the compared methods distinguish successful from unsuccessful episodes: ExpeL contrasts the two to distil insights and AWM induces workflows from episodes it treats as successful. That distinction is produced by a single language model judgement module that reads the trajectory, and every method, including \method{}, uses the same module with the same prompt. The official evaluator runs only after an episode terminates and its output reaches no method's memory. Stored experience therefore consists of observations the agent obtained by acting and of judgements formed from those observations, not of labels supplied by the benchmark, and a difference between two methods cannot arise from one of them having been given a more reliable success signal. Second, the task order is the order in which the benchmark releases its tasks; it is not chosen by us and it is identical for every method. Third, runs do not share memory, so a result for one backbone or benchmark is not carried by experience gathered in another.

Every memory based method in Table~\ref{tab:main} runs under these same conditions: the same empty start, the same task order, the same absence of benchmark labels in memory, and the same per run isolation. The methods therefore differ only in how experience is represented, selected, and injected, which is the comparison the main results are intended to support.

\subsection{Uncertainty and Claims}
The report of \MainWins{} highest means counts metric comparisons; it is not a sample of 21 independent trials or a test of significance. The $\pm$ values denote standard deviations, not standard errors or confidence intervals. Paired significance tests require outcomes on the same tasks and an explicit treatment of repeated seeds and shared task templates. The displayed marginal means and standard deviations alone do not determine confidence intervals for differences, so no statistical significance is claimed.

\section{Retrieval Procedure and Capability Management}
\label{app:implementation}
\subsection{Bounded Retrieval and Connection Procedure}
Algorithm~\ref{alg:retrieval} specifies the retrieval flow. The search optimizes the number of historical connector edges locally. Candidate endpoint operations remain part of the returned fragment even when the connector is empty because $v_a=u_b$. A path that exceeds the limit on visible operation cards is returned with an explicit truncation status or omitted as a complete connection.

\begin{algorithm}[htbp]
\caption{Retrieval of operation fragments guided by the plan}
\label{alg:retrieval}
\begin{algorithmic}[1]
\Require Task $q$, concrete history $H_t$, graph $G$, embedding function $f$, search limits $L$
\State $P_t=(S_t,D_t)\gets\operatorname{Planner}(q,H_t)$
\For{each step $s_i\in S_t$}
\State $C_i\gets$ up to three highest scoring operations matching $s_i$ above the threshold
\State Retain edge IDs, endpoints, contracts, references, and versions
\EndFor
\State $F\gets\varnothing$; initialize candidate assignments for shared steps
\For{each direct dependency $(i,j)\in D_t$ in a stable order}
\State Search $v_a\leadsto u_b$ for compatible $a\in C_i,b\in C_j$ with bounded BFS
\State Rank admissible paths by length, candidate similarity, then stable IDs
\If{a consistent connection is available within the budget}
\State Add $a$, its connector, and $b$ to $F$; record candidate assignments
\Else
\State Retain separate candidates; record gap, conflict, or search truncation
\EndIf
\EndFor
\State Expose $P_t$, $F$, unmatched candidates, and all unbound prerequisites to the main agent
\State \Return operation cards and statuses, subject to the budget for visible context
\end{algorithmic}
\end{algorithm}

With at most $k=3$ candidates per step, each direct plan dependency induces at most $k^2=9$ pair searches. If each search expands at most $B$ states, the expansion budget is at most $9|D_t|B$; actual runtime also depends on examined adjacency lists and embedding retrieval. This is a budget bound, not a claim that all retrieved fragments have globally optimal joint cost. For a failed bounded search we record whether the reachable subgraph was exhausted or the search stopped at an unexhausted frontier.

\begin{table}[htbp]
\centering\small
\begin{tabular}{lr}
\toprule
Reference parameter & Default value \\
\midrule
Operation retrieval cosine threshold / maximum candidates & 0.72 / 3 \\
Cosine threshold for merge candidates / maximum candidates & 0.82 / 3 \\
BFS maximum depth / expansions & 8 / 1,000 \\
Maximum visible operation cards & 18 \\
Skill retrieval cosine threshold / maximum cards & 0.35 / 8 \\
Code retrieval cosine threshold / maximum cards & 0.35 / 4 \\
\bottomrule
\end{tabular}
\caption{Default parameters of the reference implementation. These specify the retrieval procedure and are distinct from the settings of individual experimental runs.}
\label{tab:defaults}
\end{table}

\subsection{Execution Evidence, Capabilities, and Updates}
The main agent can use graph operations corresponding to \texttt{open}, \texttt{neighbors}, \texttt{search}, and \texttt{path}. The response format specifies direction, input requirements, observed or abstract outputs, versions, and truncation status. Retrieving a card and opening a referenced implementation are separate events. If generated code invokes a registered API or stored function whose definition has not been opened, the execution check first opens that definition and requests regenerated code. An unknown API is not created from a guessed name.

Capability counts describe stored code, skill, and role records. They do not by themselves establish automatic decomposition of arbitrary programs or learning of role prompts. The retrieval formulation assumes stable identifiers for plan steps across rounds so that dependencies and candidate assignments remain consistent.

The reflector generates concrete and abstract summaries in one response, with a further merge decision when needed. Concrete event summaries preserve actual observed values and their sources. Abstract summaries parameterize those values and retain distinctions between contracts. Validation checks placeholders against concrete evidence, verifies capability references and versions, and keeps incompatible effects separate. A memory reference indicates that an operation informed the new action; it is not evidence that a stored code template was executed unchanged. Additional retrieval and merge calls can increase the number of model calls beyond the three persistent roles, and bounded connection search does not guarantee global path optimality.

\subsection{Illustrative Dependency Example}
Consider a request to split a bill among roommates. Reading the bill produces an amount and currency; retrieving the roommate list produces recipients. Computing a split requires \emph{both} outputs. A historical path from the operation that reads the bill to the computation does not satisfy the recipient prerequisite. If another fragment supplies the roommate list, the two fragments must still refer to compatible entities and the same current task instance. Finally, requesting payments and sending money have different effects and remain distinct operations. This example illustrates contract checks and AND dependencies; it is not presented as an observed benchmark trajectory.

\section{Per Backbone Category Performance}
\label{app:radar}
\begin{figure}[tbp]
\centering\includegraphics[width=\linewidth]{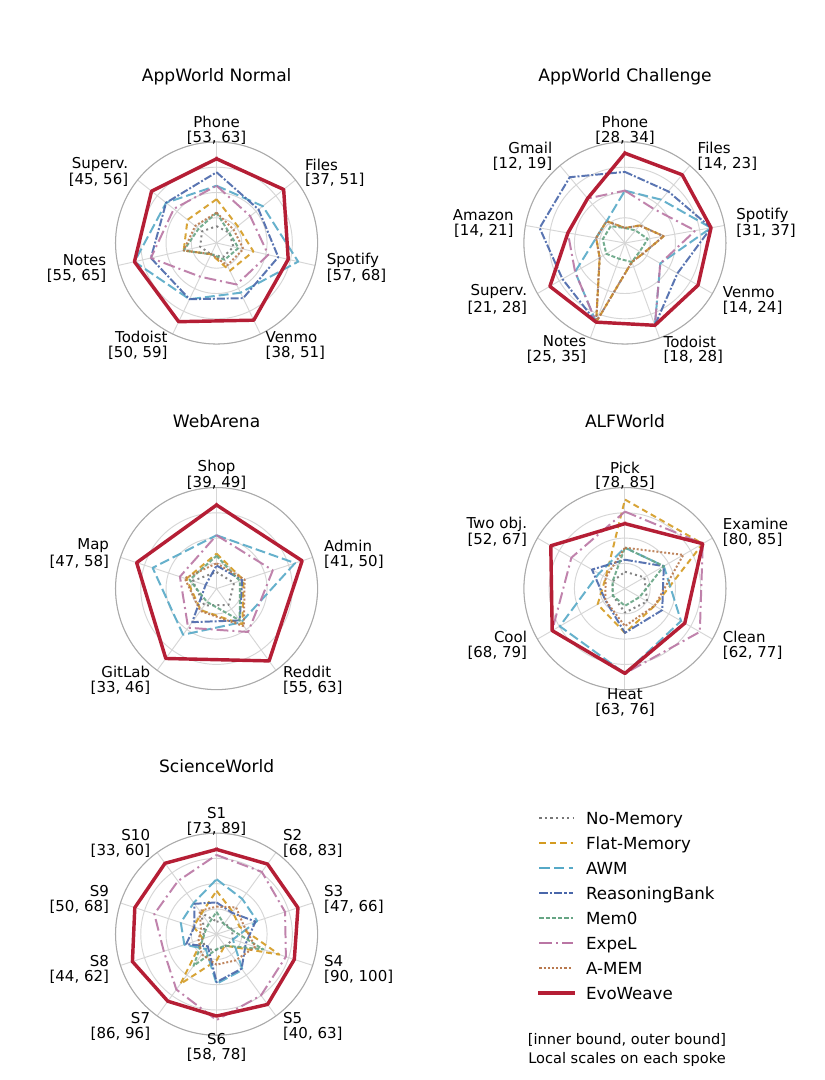}
\caption{DeepSeek-V3.2 category performance across five settings. Each spoke uses the local range printed beside its category. All eight methods are retained.}
\label{fig:radar-deepseek}
\end{figure}

\begin{figure}[tbp]
\centering\includegraphics[width=\linewidth]{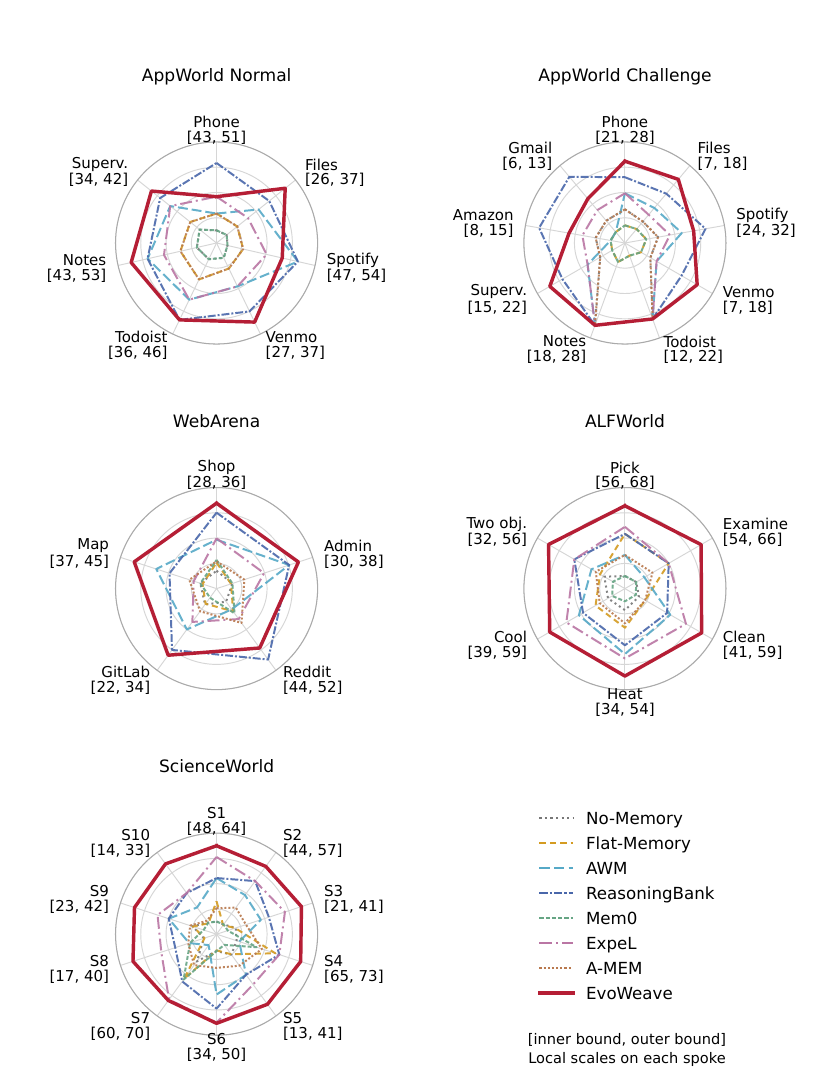}
\caption{GPT-4.1-mini category performance across the five settings. These curves are inputs from individual backbones to the equal averages over backbones in Figure~\ref{fig:radars}.}
\label{fig:radar-gpt}
\end{figure}

\begin{figure}[tbp]
\centering\includegraphics[width=\linewidth]{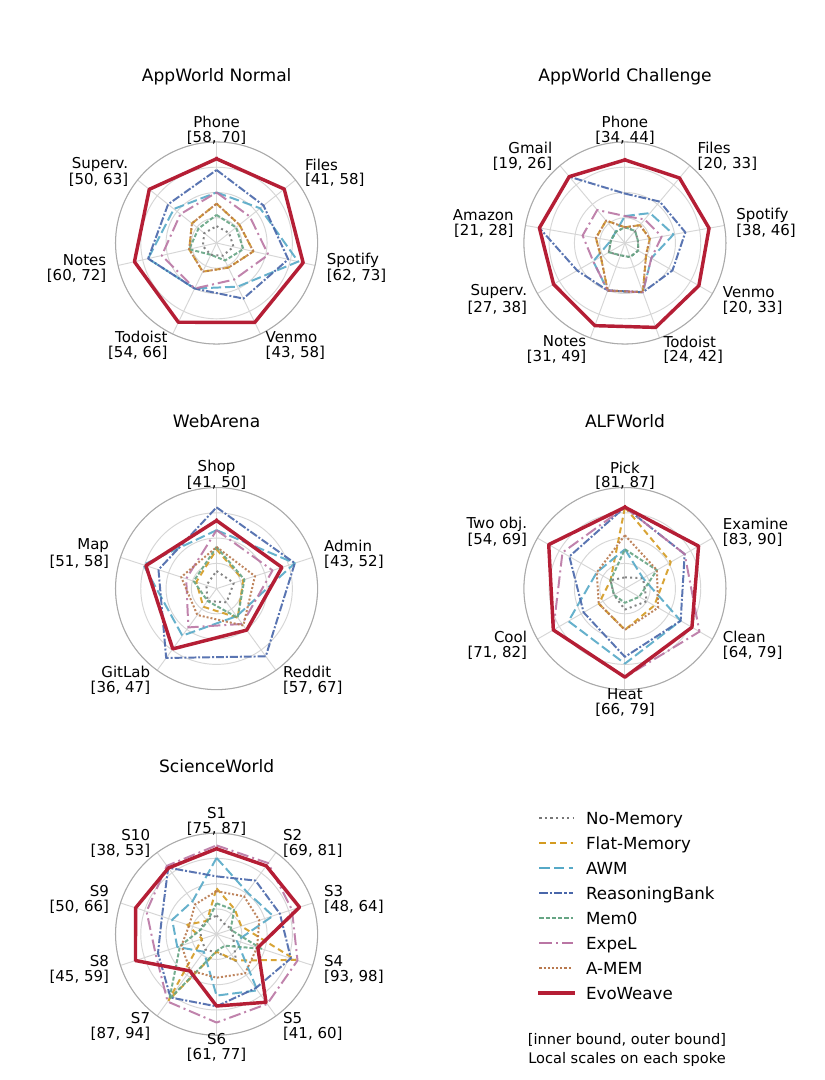}
\caption{Qwen3.8-27B category performance across the five settings. Each spoke uses the local score range printed beside its category.}
\label{fig:radar-qwen}
\end{figure}

Figures~\ref{fig:radar-deepseek} to~\ref{fig:radar-qwen} present 15 radar plots, with five settings for each backbone. Every plot includes all eight methods and all categories. A local linear scale maps each category's scores to radial positions, using the same bounds for all methods on that spoke.

\paragraph{Local axis ranges.}
For a spoke with minimum $a$ and maximum $b$ over the eight methods, let $p=\max(0.15(b-a),0.5)$. We round $a-p$ downward and $b+p$ upward, using increments of 1 when $b-a\geq3$ and 0.5 otherwise, and clip the resulting bounds $L,U$ to the valid score range from 0 to 100. The radius is $r=(x-L)/(U-L)$. All methods on that spoke share these bounds, printed as $[L,U]$ beside its label. The center denotes $L$, the outside circle denotes $U$, and intermediate rings divide this interval into quarters. Ranges are computed separately for each plotted population. Consequently, compare methods within a spoke; equal radii across spokes or backbones need not represent equal scores, and polygon area is not an aggregate performance measure.

\paragraph{Label key.}
Files denotes File\_System, Notes denotes SimpleNote, Superv. denotes Supervisor, Shop denotes Shopping, and Admin in WebArena denotes Shopping Admin. ALFWorld labels are Pick and Place, Examine in Light, Clean and Place, Heat and Place, Cool and Place, and Use Two Objects. The ScienceWorld mapping is S1: Changes in the state of matter; S2: Temperature and Melting Point; S3: Electricity and Electrical Conductivity; S4: Living and Non-living Things; S5: Plant growth; S6: Chemical mixing; S7: Animal Lifespan; S8: life cycle; S9: Physics Mechanics; S10: Mendelian genetics. Supervisor is displayed as a distinct score, with its relationship to main TGC discussed in Appendix~\ref{app:verification}.


\section{Ablation Results for Each Backbone}
\label{app:ablation}
Tables~\ref{tab:ab-deepseek} to~\ref{tab:ab-qwen} preserve the complete ablation values. For each configuration and backbone $b$, the reported Overall TGC is consistent with
\begin{equation}
\operatorname{Overall}_b=\frac{168\operatorname{Normal}_b+417\operatorname{Challenge}_b}{585}.
\end{equation}
The main text table then takes an equal mean over the three backbones. These ablations report point estimates. F is the full system, that is D with active graph retrieval; E is C with active retrieval but without directed historical connections. The highest value in each column is shown in bold. The configurations share the memory construction protocol of Appendix~\ref{app:protocol}, the same candidate pool, and the same capability library, as described in Section~\ref{sec:experiments}.

\begin{table}[htbp]\centering\small
\caption{DeepSeek-V3.2 AppWorld ablation.}\label{tab:ab-deepseek}
\begin{tabular}{lrrr}
\toprule
Configuration & Normal TGC & Challenge TGC & Overall TGC \\
\midrule
A: No memory & 47.02 & 23.02 & 29.91 \\
B: Whole workflow RAG & 49.05 & 23.02 & 30.50 \\
C: Operation cards & 49.09 & 24.36 & 31.46 \\
D: C + directed connections & 52.75 & 25.21 & 33.12 \\
E: C + active retrieval & 52.20 & 25.67 & 33.29 \\
F: EvoWeave (full) & \textbf{54.05} & \textbf{27.00} & \textbf{34.77} \\
\bottomrule
\end{tabular}

\end{table}
\begin{table}[htbp]\centering\small
\caption{GPT-4.1-mini AppWorld ablation.}\label{tab:ab-gpt}
\begin{tabular}{lrrr}
\toprule
Configuration & Normal TGC & Challenge TGC & Overall TGC \\
\midrule
A: No memory & 35.71 & 16.98 & 22.36 \\
B: Whole workflow RAG & 36.67 & 16.02 & 21.95 \\
C: Operation cards & 38.50 & 17.81 & 23.75 \\
D: C + directed connections & 39.84 & 19.80 & 25.56 \\
E: C + active retrieval & 39.11 & 18.86 & 24.68 \\
F: EvoWeave (full) & \textbf{40.60} & \textbf{21.01} & \textbf{26.64} \\
\bottomrule
\end{tabular}

\end{table}
\begin{table}[htbp]\centering\small
\caption{Qwen3.8-27B AppWorld ablation.}\label{tab:ab-qwen}
\begin{tabular}{lrrr}
\toprule
Configuration & Normal TGC & Challenge TGC & Overall TGC \\
\midrule
A: No memory & 52.02 & 29.02 & 35.63 \\
B: Whole workflow RAG & 54.05 & 30.02 & 36.92 \\
C: Operation cards & 56.20 & 31.99 & 38.94 \\
D: C + directed connections & 59.63 & 34.51 & 41.72 \\
E: C + active retrieval & 58.40 & 33.30 & 40.51 \\
F: EvoWeave (full) & \textbf{61.07} & \textbf{35.97} & \textbf{43.18} \\
\bottomrule
\end{tabular}

\end{table}

\section{Graph Statistics by Evaluation Setting}
\label{app:graph}
\begin{table}[htbp]
\centering\small\setlength{\tabcolsep}{3pt}
\caption{Reported graph statistics for each setting. Ops. denotes directed operation edges. Links denotes references from capabilities to operations, not state transition edges.}
\label{tab:graph-all}
\begin{tabular}{lrrrrrrrr}
\toprule
& States & Ops. & \multicolumn{2}{c}{Code} & \multicolumn{2}{c}{Skills} & \multicolumn{2}{c}{Roles} \\
Setting & & & Nodes & Links & Nodes & Links & Nodes & Links \\
\midrule
AppWorld Normal & 37 & 173 & 405 & 173 & 56 & 142 & 9 & 117 \\
AppWorld Challenge & 51 & 431 & 467 & 431 & 139 & 825 & 9 & 216 \\
WebArena & 18 & 67 & 17 & 67 & 180 & 562 & 5 & 61 \\
ALFWorld & 9 & 14 & 11 & 14 & 6 & 27 & 1 & 8 \\
ScienceWorld & 20 & 103 & 26 & 103 & 30 & 254 & 10 & 95 \\
\bottomrule
\end{tabular}

\end{table}
The macro means in Table~\ref{tab:graph} weight each setting equally, including Normal and Challenge as separate settings. The mean state and operation counts are \GraphStates{} and \GraphOperations{}. Mean counts of code, skill, and role nodes are \GraphCodeNodes{}, \GraphSkillNodes{}, and \GraphRoleNodes{}; the corresponding mean counts of capability links are \GraphCodeLinks{}, \GraphSkillLinks{}, and \GraphRoleLinks{}. These are arithmetic summaries of five graph snapshots. They do not identify average paths, motifs, reachability, degree distributions, or provenance quality. Such analyses require exported adjacency lists and the exact snapshot associated with each run.

\begin{figure}[tbp]
\centering\includegraphics[width=\linewidth]{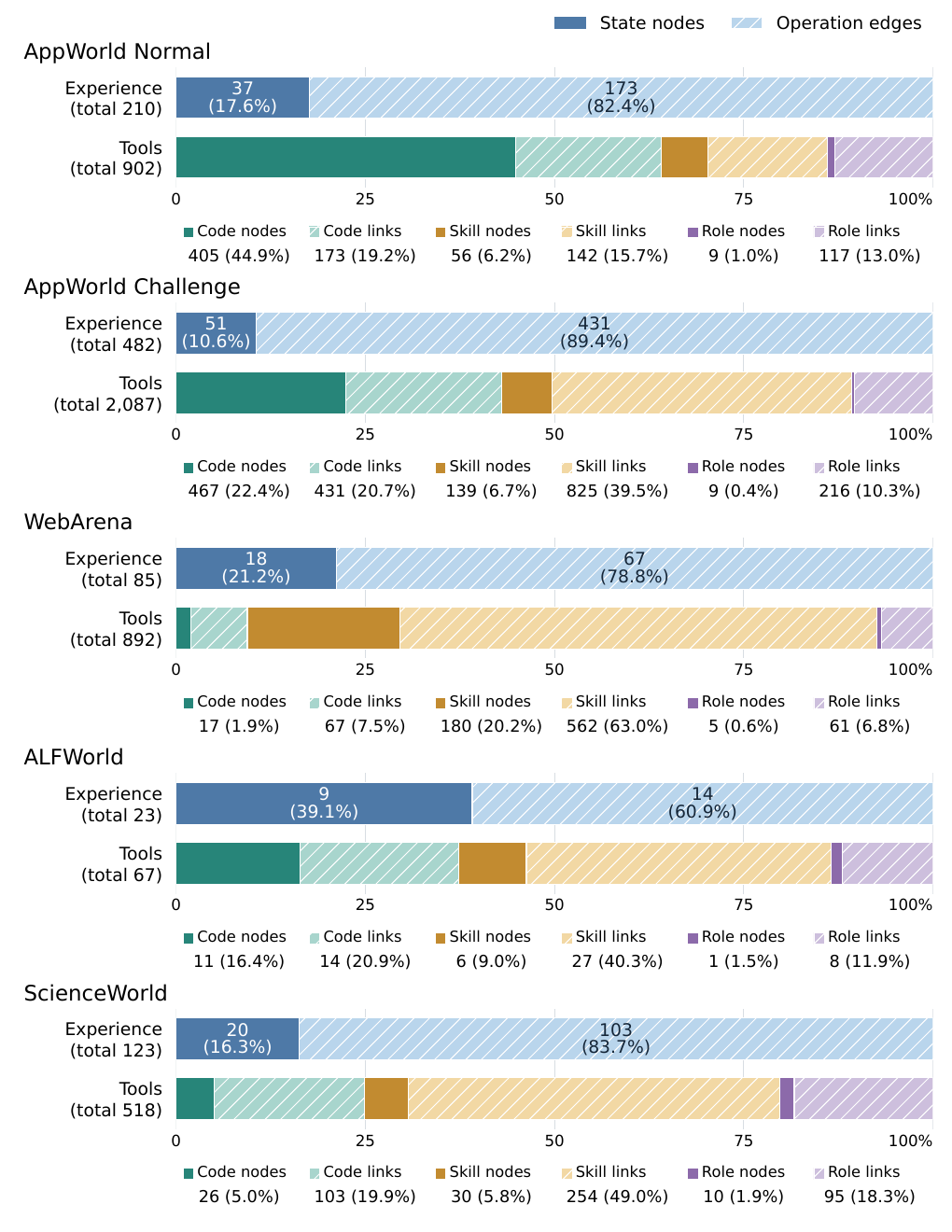}
\caption{Graph composition for the five evaluation settings. Each panel shows the experience layer and tool layer, each normalized to 100\%. Labels report counts and shares within the corresponding layer; totals appear beside the bars. Solid fills indicate nodes and hatched fills indicate operation edges or capability links. The keys below each panel give all six tool components, including those with very small shares. Source counts are retained in Table~\ref{tab:graph-all}.}
\label{fig:graph-composition}
\end{figure}

Figure~\ref{fig:graph-composition} visualizes the composition of each snapshot with two horizontal bars. The experience layer combines state nodes and operation edges. The tool layer, termed the capability layer in the method, contains code, skill, and role nodes together with their links to operations. Each bar is normalized separately: a component with count $n_k$ has share $100n_k/\sum_j n_j$, where the sum includes the displayed components of that layer. The labels retain absolute counts so that equal bar lengths do not imply equal graph sizes. These shares describe the composition of recorded graph elements, rather than memory usage or execution frequency.

\section{Token Usage and Runtime Metrics}
\label{app:efficiency}
\begin{table}[htbp]
\centering\small\setlength{\tabcolsep}{3pt}
\caption{Efficiency metrics. Injected, Input, and Output denote token counts per question. Rounds and seconds per round are averages. The summaries do not fully specify the model calls included in each token metric.}
\label{tab:efficiency}
\begin{tabular}{lrrrrrr}
\toprule
Method & Injected & Input & KV reuse (\%) & Output & Rounds & s/round \\
\midrule
No-Memory & 0.00 & 103,690.00 & 70 & 4,925.00 & 18.52 & 9.52 \\
Flat-Memory & 26,300.80 & 127,710.80 & 78 & 4,829.20 & 17.96 & 9.79 \\
AWM & 19,890.00 & 114,140.00 & 78 & 4,479.80 & 16.78 & 9.66 \\
ReasoningBank & 11,872.00 & 103,192.00 & 77 & 4,341.00 & 16.32 & 9.54 \\
Mem0 & 8,500.00 & 113,140.00 & 76 & 4,974.00 & 18.66 & 9.51 \\
ExpeL & 28,011.60 & 123,961.60 & 78 & 4,564.20 & 17.02 & 9.82 \\
A-MEM & 16,490.00 & 118,160.00 & 70 & 4,839.20 & 18.04 & 9.88 \\
EvoWeave & 6,155.46 & 102,665.46 & 45 & 4,574.00 & 15.94 & 10.44 \\
\bottomrule
\end{tabular}

\end{table}
For a memory baseline injecting $I_b$ tokens per question, the injection saving is $100(1-I_{\mathrm{\method{}}}/I_b)$. This yields \SaveFlat{}\% for Flat-Memory, \SaveAWM{}\% for AWM, and \SaveRB{}\% for ReasoningBank. The reduction in total input relative to No-Memory is $100(1-102665.46/103690)=\EnterSaving{}\%$. Injected memory tokens are zero for No-Memory, so a relative injection saving against that baseline is undefined. Lower KV reuse and more seconds per round qualify any interpretation of token savings as a runtime improvement.

Mean total latency cannot be recovered by multiplying average round count by average seconds per round: the two quantities can covary across tasks and exclude different sources of overhead. The summaries also do not establish a common accounting of reflection, validation, active search, retries, and offline construction. API monetary cost and local GPU usage require different accounting. These metrics therefore characterize context and execution tradeoffs without establishing a ranking by total latency or monetary cost.

\section{Category Operation Overlap and Referencing}
\label{app:overlap}
Let $O_i$ denote the operation set associated with task $i$ and $\mathcal T_{d,c}$ the tasks of type $c$ in setting $d$. Within a type, overlap is summarized by the mean pairwise Jaccard similarity
\begin{equation}
J_{d,c}=\frac{100}{\binom{|\mathcal T_{d,c}|}{2}}\sum_{i<j,\ i,j\in\mathcal T_{d,c}}\frac{|O_i\cap O_j|}{|O_i\cup O_j|},
\end{equation}
using benchmark specific operation proxies: gold database model sets for AppWorld, evaluation operators for WebArena, high level actions for ALFWorld, and action verbs for ScienceWorld. These sets omit ordering, arguments, and repeat counts. Overlap therefore describes shared vocabulary, with executable reuse additionally requiring compatible inputs and preconditions.

For Figure~\ref{fig:overlap}, we average the type specific percentages with equal weight. With $K_d$ types and reference percentage $R_{d,c}$,
\begin{equation}
\overline J_d=\frac{1}{K_d}\sum_{c=1}^{K_d}J_{d,c},\qquad
\overline R_d=\frac{1}{K_d}\sum_{c=1}^{K_d}R_{d,c}.
\end{equation}
The type counts are 9, 9, 5, 6, and 10 for AppWorld Normal, AppWorld Challenge, WebArena, ALFWorld, and ScienceWorld. Overlap is computed within each type, excluding cross type pairs. The averages give equal weight to types, without weighting by task counts or averaging across backbones. Memory reference percentages are analyzed separately from operation overlap. Their event definition and denominator are not sufficiently specified to interpret them as execution success rates.

\begin{figure}[htbp]
\centering\includegraphics[width=.80\linewidth]{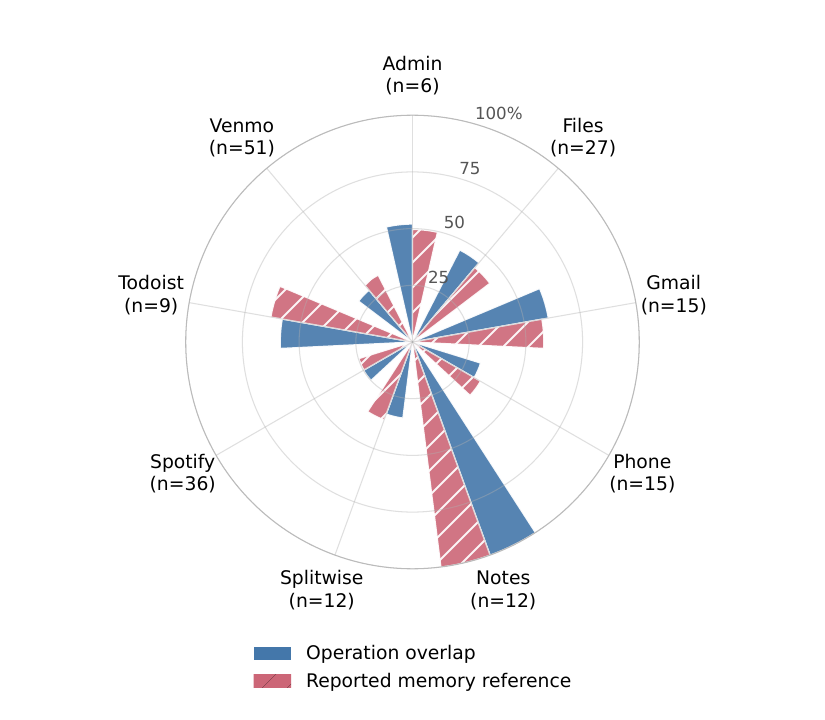}
\caption{AppWorld Normal: category overlap and reported memory referencing. Categories reflect proxies based on database models; their task counts need not partition the overall task set.}
\label{fig:overlap-normal}
\end{figure}

\begin{figure}[htbp]
\centering\includegraphics[width=.78\linewidth]{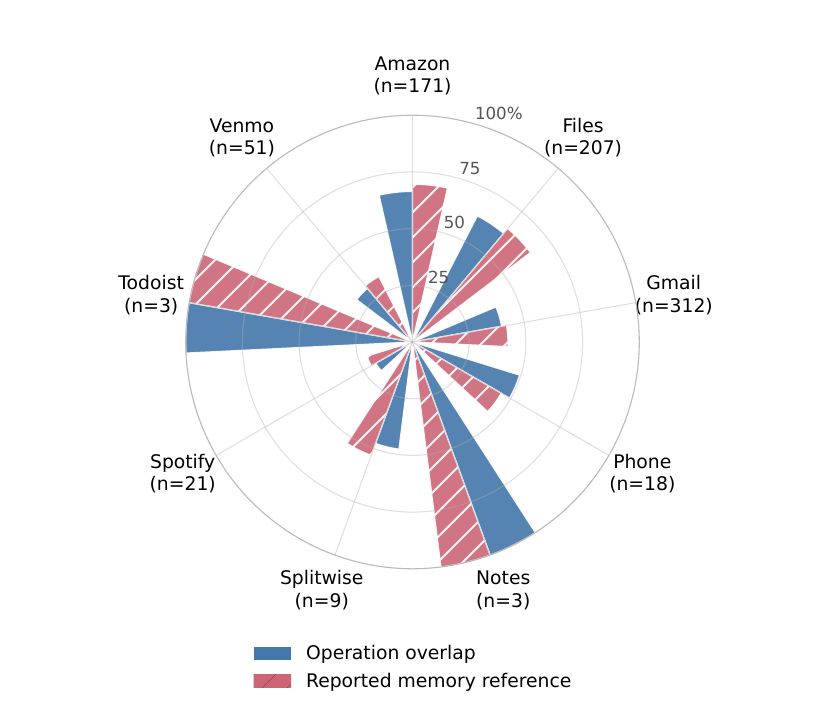}
\caption{AppWorld Challenge: category percentages, with task counts shown alongside labels. The nine types receive equal weight in Figure~\ref{fig:overlap}. Task membership can overlap across categories.}
\label{fig:overlap-challenge}
\end{figure}

\begin{figure}[htbp]
\centering\includegraphics[width=.78\linewidth]{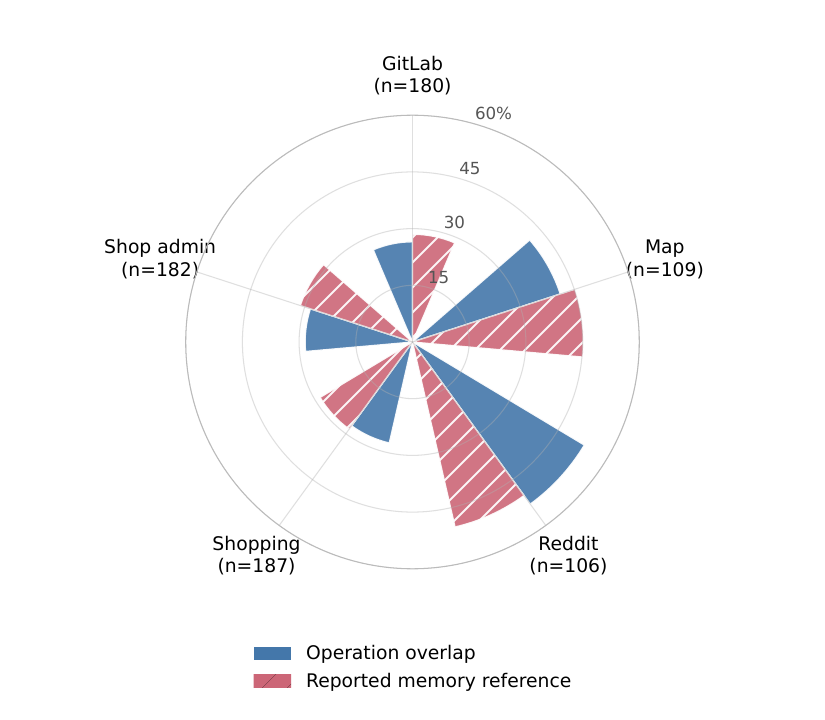}
\caption{WebArena: overlap of evaluation operators and reported memory referencing. The five single site categories cover 764 of the 812 tasks; the remaining 48 span more than one site and are included in the main table but not in any category, as explained in Appendix~\ref{app:verification}.}
\label{fig:overlap-web}
\end{figure}

\begin{figure}[htbp]
\centering\includegraphics[width=.78\linewidth]{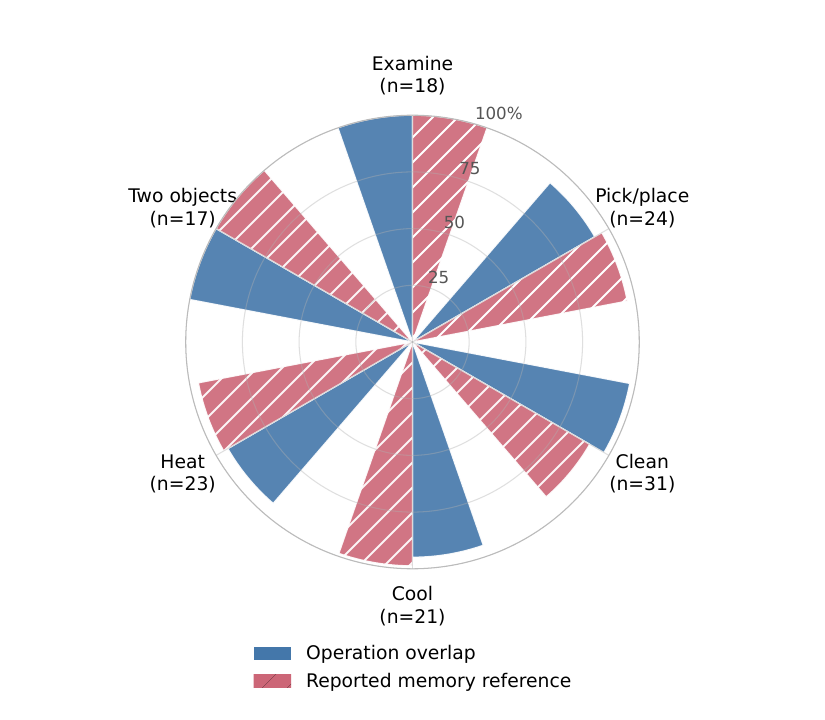}
\caption{ALFWorld: overlap of high level action sets and reported memory referencing. Repeated action vocabulary does not itself establish matching objects, preconditions, or successful execution.}
\label{fig:overlap-alf}
\end{figure}

\begin{figure}[htbp]
\centering\includegraphics[width=.78\linewidth]{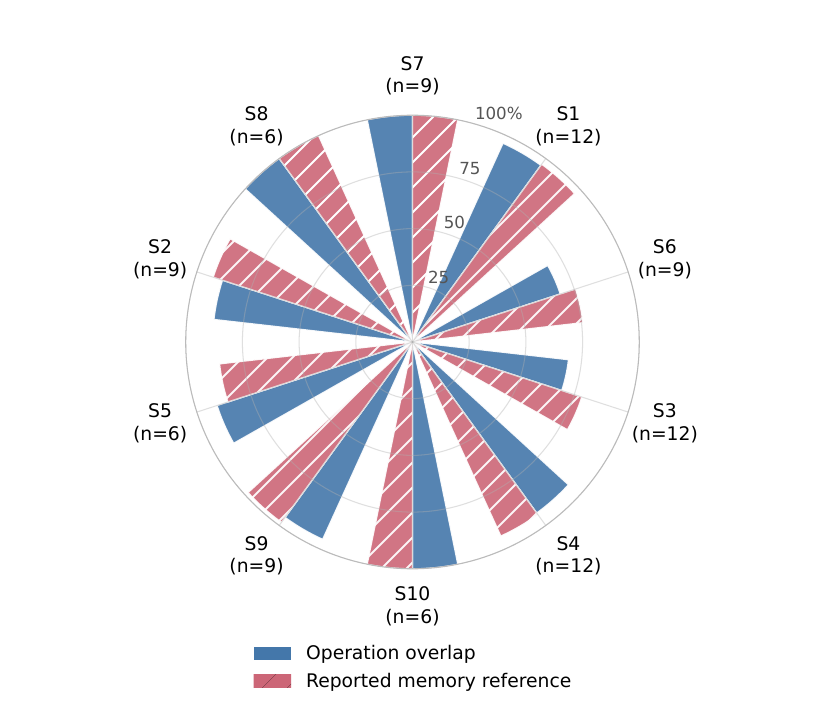}
\caption{ScienceWorld: overlap of gold verb sets and reported memory referencing across 90 tasks. S1 to S10 follow the topic key in Appendix~\ref{app:radar}.}
\label{fig:overlap-science}
\end{figure}

Figures~\ref{fig:overlap-normal} to~\ref{fig:overlap-science} show all 39 types and their task counts. AppWorld memberships can overlap. ScienceWorld's auxiliary sample covers 90 tasks from 30 templates with three instances each, grouped into ten topics; complete benchmark coverage is not established. Perfect overlap in small types should be interpreted in light of the size of the underlying vocabulary and of the sample.


\clearpage
\section{Further Evaluation Directions}
\label{app:further}
The existing results motivate, but do not replace, stronger mechanism tests. A token matched comparison of C and D should preserve retrieved operation text and capability references while controlling the number of visible tokens; otherwise, the graph variant can benefit simply from more context. F should receive the same budget for searching descriptions and opening bodies as E, with adjacency and path information as the intended difference. A comparison with fixed capabilities should freeze API definitions and code functions across configurations, isolating changes to experience organization.

A composition test should distinguish no overlap, overlap in one step, partial overlap in several steps, novel combinations of known operations, and requirements with AND joins. The task split must be determined without selecting on final success. Retrieval sensitivity should report both behavior when there are no matches and the effect of thresholds, candidate counts, path depth, and expansion limits. Candidate conflicts, truncated searches, and missing prerequisites should be logged rather than absorbed into a generic failure rate.

A frozen memory condition would complement the online protocol of Appendix~\ref{app:protocol} by building memory on a separate split and then evaluating with it fixed, which isolates transfer from continual accumulation. Reporting accuracy as a function of position in the evaluation stream would characterize how quickly useful experience is acquired. Human review of a sample of reused operations could measure whether a reference was relevant, correctly parameterized, and actually adopted. These directions extend the present analyses; they are not additional experimental results.

\end{document}